\documentclass[letterpaper]{article} 
\usepackage{aaai2026}  
\usepackage{times}  
\usepackage{helvet}  
\usepackage{courier}  
\usepackage[hyphens]{url}  
\usepackage{graphicx} 
\usepackage{natbib}  
\usepackage{caption} 
\usepackage{algorithm}
\usepackage{algorithmic}
\usepackage{amsmath}
\usepackage{newfloat}
\usepackage{listings}
\DeclareCaptionStyle{ruled}{labelfont=normalfont,labelsep=colon,strut=off} 
\floatstyle{ruled}
\newfloat{listing}{tb}{lst}{}
\floatname{listing}{Listing}
\usepackage{amsmath}
\usepackage{amssymb}
\usepackage{mathtools}
\usepackage{amsthm}
\usepackage{enumitem} 

\theoremstyle{plain}
\newtheorem{theorem}{Theorem}[section]

\theoremstyle{definition}
\newtheorem{definition}[theorem]{Definition}

\theoremstyle{remark}

\usepackage{multirow} 
\usepackage{makecell}
\usepackage{amsthm} 
\newtheorem{problem}{Problem} 
\usepackage[most]{tcolorbox} 
\tcbuselibrary{breakable} 
\usepackage{booktabs}
\usepackage{courier} 
\newcommand{\constant}[1]{\text{\texttt{#1}}}

\newcommand{\StateCommentLabel}[3]{
	\STATE #1 \label{#3} \hfill $\triangleright$ #2
}
\newcommand{\StateComment}[2]{
	\STATE #1 \hfill $\triangleright$ #2
}

\newcommand{\State}[1]{
	\STATE #1
}

\newlength\myindent
\newcommand\tuple[1]{\left<#1\right>}

\title{CABTO: Context-Aware Behavior Tree Grounding for Robot Manipulation}
\author{
    Yishuai Cai\textsuperscript{\rm 1,\rm 2, \rm 4}\equalcontrib\thanks{This work was completed during an internship at PsiBot.},
    Xinglin Chen\textsuperscript{\rm 1,\rm 2, \rm 4}\equalcontrib,
    Yunxin Mao\textsuperscript{\rm 1},
    Kun Hu\textsuperscript{\rm 1},\\
    Yaodong Yang\textsuperscript{\rm 3, \rm 4}, 
    Yuanpei Chen\textsuperscript{\rm 2, \rm 3, \rm 4},
    Wenjing Yang\textsuperscript{\rm 1},
    Ji Wang\textsuperscript{\rm 1},
    Minglong Li\textsuperscript{\rm 1}\thanks{Corresponding Author.},
}
\affiliations{

    \textsuperscript{\rm 1}National University of Defense Technology\quad
    \textsuperscript{\rm 2}PsiBot\quad
    \textsuperscript{\rm 3}Peking University\quad
    \textsuperscript{\rm 4}PKU-Psibot Lab\\
    
    \{caiyishuai, chenxinglin, liminglong10\}@nudt.edu.cn
}

\usepackage{bibentry}

\begin{document}

\maketitle

\begin{abstract}
Behavior Trees (BTs) offer a powerful paradigm for designing modular and reactive robot controllers. BT planning, an emerging field, provides theoretical guarantees for the automated generation of reliable BTs. However, BT planning typically assumes that a well-designed BT system is already grounded—comprising high-level action models and low-level control policies—which often requires extensive expert knowledge and manual effort. In this paper, we formalize the BT Grounding problem: the automated construction of a complete and consistent BT system. We analyze its complexity and introduce CABTO (Context-Aware Behavior Tree grOunding), the first framework to efficiently solve this challenge. CABTO leverages pre-trained Large Models (LMs) to heuristically search the space of action models and control policies, guided by contextual feedback from BT planners and environmental observations. Experiments spanning seven task sets across three distinct robotic manipulation scenarios demonstrate CABTO’s effectiveness and efficiency in generating complete and consistent behavior tree systems.
\end{abstract}

\begin{links}
    \link{Code}{https://github.com/DIDS-EI/CABTO}
\end{links}

\section{Introduction}

Robot manipulation necessitates both reliable high-level planning and robust low-level control policies. Recently, Behavior Trees (BTs) \cite{ogren2022behavior,colledanchise2018behavior} have emerged as a highly reliable and robust control architecture for intelligent robots, recognized for their modularity, interpretability, reactivity, and safety. Many methods have been proposed to automatically generate BTs for task execution, including evolutionary computing \cite{neupane2019learning,colledanchise2019learning} and machine learning approaches \cite{banerjee2018autonomous,french2019learning}. In particular, BT planning \cite{cai2021bt,chen2024integrating,cai2025mrbtp} has shown significant promise, primarily due to its strong theoretical guarantees: the BTs generated by such methods are provably successful in achieving goals within a finite time horizon.


Despite these advancements, BT planning critically assumes the \textit{prior existence} of a well-grounded BT system. Constructing such a system, encompassing both high-level action models and their corresponding low-level control policies, typically requires substantial human expertise and effort. Specifically, for high-level planning, the BT system must contain a sufficient and appropriately modeled set of condition and action nodes, enabling their assembly into BTs capable of accomplishing diverse tasks. Concurrently, for low-level execution, these nodes must be reliably linked to executable control policies that ensure environmental transitions occur precisely as specified by the action models, ideally with high success rates.

\begin{figure*}[t]
    \centering
    \includegraphics[width=0.98\textwidth]{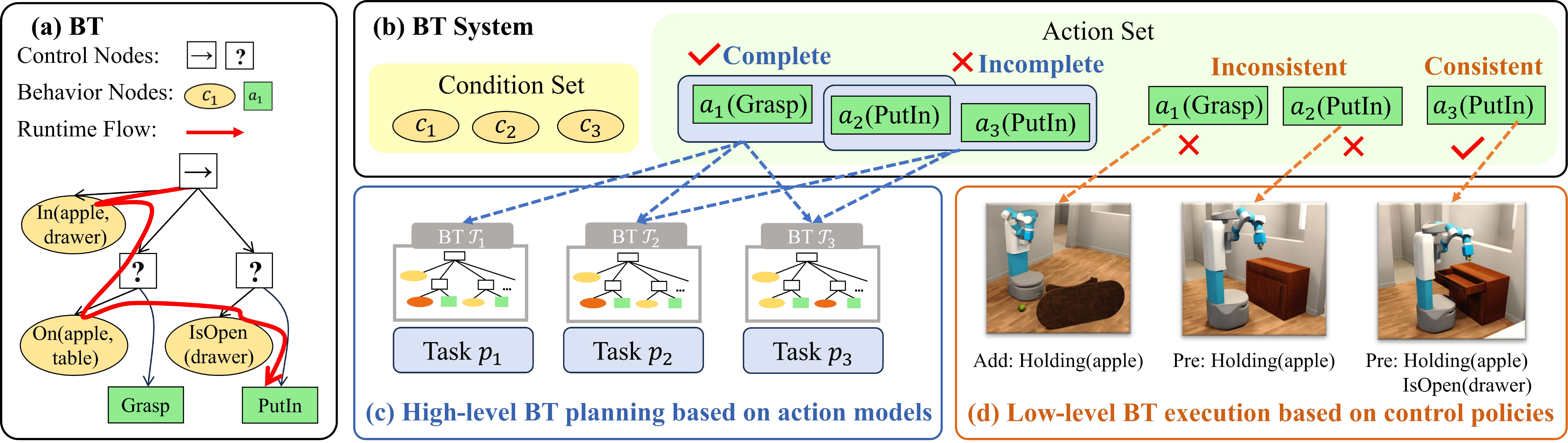}
    \caption{Concepts involved in the BT grounding problem. (a) A BT is a directed rooted tree with behavior nodes and control nodes. (b) The solution is a complete and consistent BT system for the given task set. (c) A complete BT system can generate solution BTs for all tasks during the high-level BT planning based on action models. (d) A consistent BT system ensures its control policies result in state transitions consistent with their action models during low-level BT execution.}
    \label{fig:introduction}
\end{figure*}

In this paper, we formally define the BT grounding problem: the automated construction of a complete and consistent BT system for a given task set. We characterize a well-designed BT system by two critical properties: (1) Completeness: a complete BT system can generate solution BTs for all tasks within the specified task set through high-level BT planning, based on its action models. (2) Consistency: a consistent BT system ensures that its control policies lead to state transitions that precisely match their corresponding action models during low-level BT execution. Figure \ref{fig:introduction} illustrates these concepts. For instance, a BT system with action set $\left\{a_2,a_3\right\}$ is incomplete if it can only produce a solution BT for a subset of tasks, such as $\left\{p_2,p_3\right\}$. Conversely, an action set like $\left\{a_1,a_2\right\}$ that successfully generates solution BTs for all three tasks exemplifies completeness. However, $a_1$ would be inconsistent if its control policy fails to achieve $\mathtt{Holding(apple)}$ as declared by its action model. Furthermore, $a_2$ is also inconsistent because its policy cannot put the apple in the drawer without the precondition $\mathtt{IsOpen(drawer)}$. In contrast, an action like $a_3$, whose policy perfectly aligns with its action model's state transitions, is considered consistent.

We demonstrate a naive algorithm that solves the BT grounding problem via exhaustive search. While this approach effectively illustrates the core concepts, its exponential time complexity renders it impractical for deployment.

Large Models (LMs), pre-trained on extensive corpora, images or datasets in other modalities, have shown significant abilities in searching, reasoning and planning \cite{zhou2024language,valmeekam2023planbench,HBTP}. Leveraging the advantages of LMs, we propose the first framework for efficiently solving the BT grounding problem, named Context-Aware Behavior Tree grOunding (CABTO). CABTO mainly utilizes pre-trained LMs to heuristically search the space of action models and control policies based on the contexts of BT planning and environmental feedback. 

CABTO includes three phases: (1) High-level model proposal. Given a task set, we first use Large Language Models (LLMs) to generate promising action models and use a sound and complete BT planning algorithm to evaluate their completeness. The contexts here in this phase are planning details. (2) Low-level policy sampling. We then employ Vision-Language Models (VLMs) to sample promising policy types as well as their hyperparameters for explored action models. The matching policy and its corresponding action model together form a consistent action. The contexts in this phase are environment feedbacks. (3) Cross-level refinement. If the algorithm fails to find any policy for a given action model, then it is determined to be inconsistent. In this case, the contexts of both high-level planning information and low-level environment feedbacks can be combined to help VLMs to refine the action model and generate more promising action models. 

The key contributions of this work are as follows:

\begin{itemize}
    \item We formally define the BT grounding problem as the construction of a complete and consistent BT system for a given task set. We provide a formal analysis and present a naive algorithm that elucidates the foundational concepts for solving this problem.
    \item  We propose CABTO, the first framework for efficiently solving the BT grounding problem. CABTO strategically utilizes pre-trained LMs to heuristically explore the space of action models and control policies, informed by both BT planning contexts and environmental feedback.
    \item We empirically validate CABTO's superior effectiveness and efficiency in automatically generating complete and consistent BT systems across 7 diverse task sets in 3 distinct robotic manipulation scenarios. Comprehensive ablation studies further investigate the impact of LMs, control policy types, and cross-level refinement.
\end{itemize}

\section{Related Work}
\subsubsection{Behavior Tree Generation}

Most existing BT generation methods focus on constructing the BT structure while assuming predefined execution policies. Heuristic search approaches, including grammatical evolution \cite{neupane2018geese}, genetic programming \cite{lim2010evolving} , and Monte Carlo DAG Search \cite{scheide2021behavior}, have been widely studied. Machine learning methods, such as reinforcement learning \cite{banerjee2018autonomous,pereira2015framework}and imitation learning \cite{french2019learning}, as well as formal synthesis approaches like LTL \cite{neupane2023designing} and its variants\cite{tadewos2022specificationguided}, have also been explored. However, these methods often require complex environment modeling or cannot guarantee BT reliability. In contrast, BT planning \cite{HBTP,chen2024integrating,cai2025mrbtp}  based on STRIPS-style action models \cite{fikes1971strips} provides interpretable environment modeling while ensuring both reliability and robustness.


\subsubsection{High-Level Action Models}
Action models define the blueprints of actions that drive state transitions in a system \cite{arora2018review}. They are widely used in classical planning \cite{hoffmann2001ff,bonet1997robust}, task and motion planning (TAMP) \cite{yang2024guiding,kumar2024openworld}, and symbolic problem solving \cite{pan2023logiclm,fikes1971strips}. To reduce expert design effort, many methods learn action models from plan execution traces \cite{mahdavi2024leveraging,bachor2024learning,mordoch2024safe,liu2023llm}, employing inductive learning \cite{liangvisualpredicator}, evolutionary algorithms \cite{newton2010implicit}, reinforcement learning \cite{rodrigues2012active}, and transfer learning \cite{zhuo2014actionmodel}. However, these approaches typically assume the traces are already available, overlooking how to obtain them through low-level execution—an obstacle to practical deployment.

\subsubsection{Low-Level Control Policies}
Modern low-level robot manipulation policies can be broadly categorized into three types: (1) End-to-end policies, which directly map proprioceptive inputs to joint controls via reinforcement learning \cite{bai2025retrieval,chen2023sequential}, imitation learning \cite{zare2024survey}, and, more recently, Vision-Language-Action Models (VLAs) fine-tuned from large vision-language models \cite{zhong2025survey,kim2024openvla,zhen20243dvla}. (2) Hierarchical policies, which decompose control into structured modules leveraging representations such as rigid-body poses \cite{kaelbling2011hierarchical}, constraints \cite{huang2024rekep}, affordances \cite{huang2023voxposer}, waypoints \cite{zhang2024pivotr}, or skills and symbolic codes \cite{haresh2024clevrskills,mu2024robocodex}. These approaches exploit expert knowledge to improve interpretability and extend long-horizon capabilities. (3) Rule-based policies, built solely on expert-designed control algorithms \cite{thomason2024motions,2023curobo}, offer strong robustness for specific tasks but struggle to generalize to unseen scenarios.

\section{Preliminaries}

\paragraph{Behavior Tree}

A BT $\mathcal{T}$ is a rooted directed tree where internal nodes are control flow nodes and leaf nodes are execution nodes \cite{colledanchise2018behavior}. The tree is executed via periodic "ticks" from the root. The core nodes include: (1) Condition: returns {\ttfamily success} if a state proposition holds, else {\ttfamily failure}. (2) Action: performs tasks and returns {\ttfamily success}, {\ttfamily failure}, or {\ttfamily running}. (3) Sequence ($\rightarrow$): succeeds only if all children succeed (AND logic). (4) Fallback ($?$): fails only if all children fail (OR logic).

\paragraph{BT System}
Following \cite{cai2021bt}, a BT can be represented as a four-tuple $\mathcal{T} = \langle n, h, \pi, r \rangle$, where $n$ is the number of binary propositions describing the world state. Here, $h: 2^n \to 2^n$ denotes the action model representing the intended state transition; $\pi: 2^n \to 2^n$ denotes the control policy representing the actual execution effect; and $r:2^{n}\mapsto \{ $\constant{success}, \constant{running}, \constant{failure}\} partitions the state space according to the BT’s return status.

A BT system is defined as $\Phi = \langle \mathcal{C}, \mathcal{A} \rangle$. Each action $a \in \mathcal{A}$ is a tuple $\langle h_a, \pi_a \rangle$, where $h_a = \langle pre^h(a), add^h(a), del^h(a) \rangle$ is its action model (intended effect) and $\pi_a = \langle pre^\pi(a), add^\pi(a), del^\pi(a) \rangle$ is its control policy (actual effect). The precondition $pre^h(a),pre^\pi(a)$, add effects $add^h(a),add^\pi(a)$, and delete effects $del^h(a),del^\pi(a)$ are all the subset of the condition node set $\mathcal{C}$. In a well-designed BT system, provided that the current state $s_t$ satisfies the precondition (i.e., $s_t \supseteq pre^h(a)$), the state transition upon completion of action $a$ after $k$ time steps satisfies:
\begin{equation}\label{eqn:s_f}
s_{t+k} = h_a(s_t) = \pi_a(s_t) = s_t\cup add(a)\setminus del(a) 
\end{equation}
where $h_a(s_t)$ and $\pi_a(s_t)$ denote the states resulting from the action model and the control policy execution, respectively.



\paragraph{BT Planning}

Given a BT system $\Phi$, a BT planning problem is defined as: \( p= \left<\mathcal{S}, s_0,g\right>\), where \( \mathcal{S}\) is the finite set of environment states, $s_0$  is the initial state, $g$ is the goal condition. A condition $c \subseteq \mathcal{C}$ is a subset of a state $s$, and can be an atom condition node or a sequence node with atom condition nodes as children. If $c\subseteq s$, then $c$ holds in $s$. A sound and complete BT planning algorithm, like BT Expansion \cite{cai2021bt}, ensures a solution BT $\mathcal{T}$ in finite time if $p$ is solvable. Such a BT $\mathcal{T}$ can transition the state from $s_0$ to $s_n =\pi_\mathcal{T}(s_0) \supseteq g$ in a finite number of steps $n$.


\begin{algorithm}[t]
\small
\caption{Naive Algorithm for BT Grounding}
\label{alg:general-solution}
\textbf{Input}: Problem $\tuple{\mathcal{P},\mathcal{C}_{\mathcal{P}}, \mathcal{H}_{\mathcal{P}},\Pi_\mathcal{P}}$ \\
\textbf{Output}: Solution $\Phi = \langle \mathcal{C}, \mathcal{A} \rangle$
\begin{algorithmic}[1]

\StateCommentLabel{$\mathcal{A} \leftarrow \emptyset$}{initialize grounded actions}{line1-init} \\

\FOR{$pre \in 2^{\mathcal{C}_{\mathcal{P}}}, add \in 2^{\mathcal{C}_{\mathcal{P}}},del \in 2^{\mathcal{C}_{\mathcal{P}}}$} \label{line1-fullhspace}

    \StateComment{$h \leftarrow\left<pre,add,del\right>$}{create an action model} \label{line1-actionmodel}
    \IF{$h \in \mathcal{H}_\mathcal{P}$} \label{line1-hspace}
        \FOR{\textbf{each} policy $\pi \in \Pi_\mathcal{P}$}  \label{line1-pispace}
            \IF{\constant{Consistent}$(h, \pi)$} \label{line1-policy}
                \StateComment{$a\leftarrow \left<h,\pi\right>$}{create a consistent action} \label{line1-action}
                \StateComment{$\mathcal{A}\leftarrow \mathcal{A}\cup\left\{a\right\}$}{add the consistent action} \label{line1-addaction}
                \State{\textbf{break}}
            \ENDIF
        \ENDFOR
    \ENDIF

\ENDFOR
\State{$\mathcal{C} \leftarrow \bigcup_{a\in\mathcal{A}} pre^h(a) \cup add^h(a) \cup del^h(a)$} \label{line1-condition}

\RETURN $\Phi=\left<\mathcal{C},\mathcal{A}\right>$ 
\label{line1-return}

\end{algorithmic}
\end{algorithm}

\begin{algorithm}[t]
\caption{CABTO}
\small
\label{alg-cabto}
\textbf{Input}: Problem $\tuple{\mathcal{P},\mathcal{C}_{\mathcal{P}}, \mathcal{H}_{\mathcal{P}},\Pi_\mathcal{P}}$ \\
\textbf{Output}: Solution  $\Phi = \langle \mathcal{C}, \mathcal{A} \rangle$
\begin{algorithmic}[1]
\StateComment{$\mathcal{A} \leftarrow \emptyset$}{initialize grounded actions} \label{line1-initaction}
\StateComment{$\mathcal{H}_U \leftarrow \mathcal{H}_{\mathcal{P}}$}{initialize model search spaces} \label{line2-initspaces}
\StateComment{$\mathcal{H}_E \leftarrow $\constant{LLM}($\mathcal{P},\mathcal{H}_{\mathcal{P}}$)}{Equation \ref{equ-initproposal}}\label{line3-initproposal}

\WHILE{$\mathcal{H}_U \neq \emptyset$ \textbf{and} \NOT \constant{AllSolvable}($\mathcal{P}$, $\mathcal{A}$)  } \label{line2-whileHU}
\State{\texttt{\hfill {// \textbf{high-level model proposal}}} }

    \REPEAT
        \STATE $\mathcal{I}_{fail} \leftarrow \{ \mathcal{I}_p \mid p \in \mathcal{P}, 
 $\constant{BTPlanning}$(p, \mathcal{H}_E) \text{ fails} \}$ \label{line4-planning}
        \IF{$\mathcal{I}_{fail} \neq \emptyset$}
            \StateComment {$\mathcal{H}' \leftarrow \text{\constant{LLM}}(\mathcal{P}, \mathcal{H}_U, \mathcal{I}_{fail})$} {Equation \ref{equ-modelproposal}} \label{line5-modelproposal}
            \STATE $\mathcal{H}_U \leftarrow \mathcal{H}_U \setminus \mathcal{H}', \mathcal{H}_E \leftarrow \mathcal{H}_E \cup \mathcal{H}'$ \label{line6-deleteh}
        \ENDIF
    \UNTIL{$\mathcal{I}_{fail} = \emptyset$ \textbf{or} $\mathcal{H}_U = \emptyset$}

    \State{\texttt{\hfill{// \textbf{low-level policy sampling}}} }
        \FOR{\textbf{each} $h \in \mathcal{H}_E \setminus \mathcal{H}$} \label{line-phase2-start}
            \STATE $n \leftarrow 0$, \quad $\pi \leftarrow \mathtt{null}$, \constant{Consistent}$(h, \pi)$ $\leftarrow$ False
            \WHILE{$n < N_{max}$ \AND \NOT \constant{Consistent}$(h, \pi)$} \label{line-nmax}

                \StateComment{$\pi\leftarrow$\constant{VLM}($h,\Pi_\mathcal{P},\mathcal{I}_e$) }{Equation \ref{equ-policysampling_v3}}  \label{line-vlm-sample}
                \STATE Sample a scenario $s_0$ where $pre(h) \subseteq s_0$ \label{line-sample-s0}
                \STATE $s_t, \mathcal{I}_e \leftarrow \text{\constant{Execute}}(\pi, s_0)$ \label{line-execute}
                
                \IF{$s_t \supseteq (pre(h) \cup add(h) \setminus del(h))$} \label{line-check-logic}
                    \STATE \constant{Consistent}$(h, \pi)$ $\leftarrow$ True
                    \STATE $\mathcal{A} \leftarrow \mathcal{A} \cup \{ \langle h, \pi \rangle \}, \mathcal{H} \leftarrow \mathcal{H} \cup \{h\}$

                \ENDIF
                \STATE $n \leftarrow n + 1$
            \ENDWHILE

    \State{\texttt{\hfill{// \textbf{cross-level refinement}}} }
    \IF  {\NOT \constant{Consistent}$(h, \pi)$}
        \StateComment {$h'  \leftarrow$ \constant{VLM}($h,\mathcal{H}_U,\left\{\mathcal{I}_p\right\},\Pi_\mathcal{P},\left\{\mathcal{I}_e\right\}$)} {Equation \ref{equ-crosslevel}} \label{line2-crosslevel} 
        \State{$\mathcal{H}_U \leftarrow \mathcal{H}_U \setminus \{h'\}, \mathcal{H}_E \leftarrow \mathcal{H}_E \cup \{h'\}$} \label{line2-delete2}
    \ENDIF
\ENDFOR

\StateComment {$\mathcal{H}_E \leftarrow \mathcal{H}$}  {Prune $\mathcal{H}_E$ to validated set} \label{line32-updateh}

\ENDWHILE
\State{$\mathcal{C} \leftarrow \bigcup_{\langle h, \pi \rangle \in \mathcal{A}} (pre(h) \cup add(h) \cup del(h))$}\label{line2-conditionset}

\RETURN $\Phi=\left<\mathcal{C},\mathcal{A}\right>$ \label{line2-return}
\end{algorithmic}
\end{algorithm}

\subsection{Problem Formulation}

In this paper, we focus on the automatic construction of the BT system, and therefore need to formally define the properties that describe a well-designed BT system.

\begin{definition}[Completeness]
A BT system $\Phi$ is complete in the task set $\mathcal{P}$ if, $\forall p \in \mathcal{P}$, any complete BT planning algorithm can produce a BT $\mathcal{T}$ that solves the task $p$ according to its action models.
\end{definition}

The completeness of the BT system $\Phi$ describes whether its condition nodes $\mathcal{C}$ and action nodes $\mathcal{A}$ are sufficient to solve all of the tasks in the given task set at the planning level. 

\begin{definition}[Consistency]
An action $a$ is consistent if $pre^\pi(a)\subseteq pre^h(a), add^\pi(a)=add^h(a),del^\pi(a)=del^h(a)$. That is, the control policy $\pi_a$ is capable of inducing state transitions that match its action model.
 A BT system $\Phi$ is consistent if $\forall a\in \mathcal{A}, a$ is consistent.
\end{definition}

The consistency of the BT system $\Phi$ describes whether all action nodes can be successfully executed and cause the state to transition as desired, just as specified by their action models. Both completeness and consistency are essential for constructing a BT system for embodied robots to complete tasks. 
We then define the BT grounding problem as follows:



\begin{problem}[BT Grounding]
A BT grounding problem is a tuple \(\left<\mathcal{P},\mathcal{C}_{\mathcal{P}},\mathcal{H}_{\mathcal{P}},\Pi_{\mathcal{P}}\right>\), where \( \mathcal{P} \) is the finite task set, $\mathcal{C}_{\mathcal{P}}$ is the finite set of valid condition nodes, $\mathcal{H}_{\mathcal{P}}$ is the finite set of valid action models, $\Pi_{\mathcal{P}}$ is the set of valid control policies.
A solution to this problem is a BT system $\Phi = \left<\mathcal{C},\mathcal{A}\right>$ that is complete and consistent in the task set $\mathcal{P}$, where $\mathcal{C}\subseteq \mathcal{C}_{\mathcal{P}}$, $\forall a\in\mathcal{A}, a=\left<h_a,\pi_a\right>, h_a\in\mathcal{H}\subseteq \mathcal{H}_\mathcal{P},\pi_a\in\Pi\subseteq \Pi_\mathcal{P}$.
\end{problem}

\begin{figure*}
	\centering
	\includegraphics[width=\textwidth]{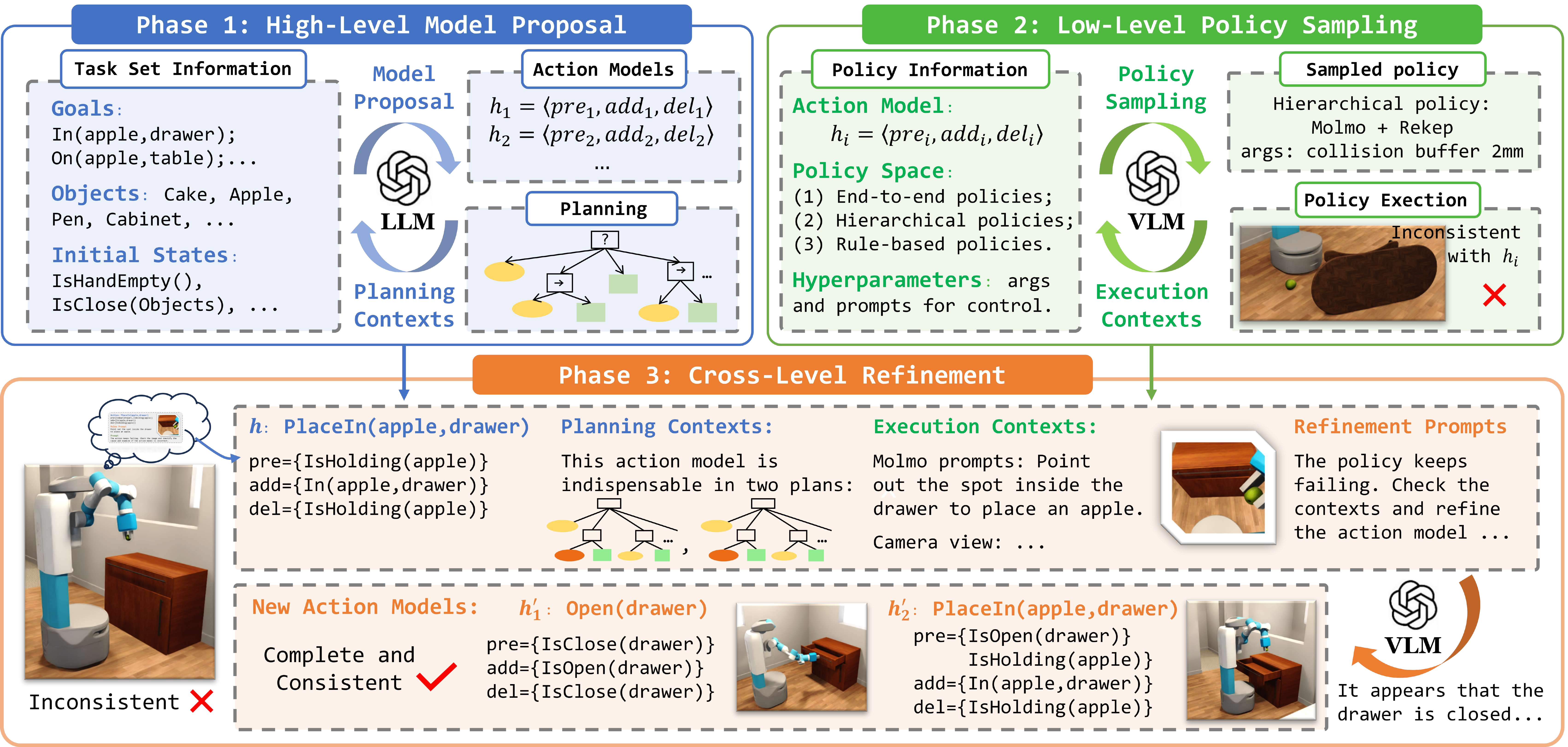}

	\caption{The framework of CABTO includes three phases: (1) High-level model proposal leverages the planning contexts for the LLMs to heuristically explore the space of action models; (2) Low-level policy sampling leverages the execution contexts for the VLMs to heuristically explore the space of control policies; (3) Cross-level refinement leverages both planning and execution contexts for refining inconsistent action models.}

	\label{fig:framework}
\end{figure*}

\section{Methodology}

This section first presents a naive algorithm and a formal analysis to establish the foundational principles of the BT grounding problem. We then detail the CABTO framework, encompassing high-level model proposal, low-level policy sampling, and cross-level refinement. Finally, we provide the implementation details of the CABTO system.

\subsubsection{Naive Algorithm for BT Grounding}
Algorithm \ref{alg:general-solution} outlines a naive approach to BT grounding. Given the problem tuple $\tuple{\mathcal{P},\mathcal{C}_{\mathcal{P}}, \mathcal{H}_{\mathcal{P}},\Pi_\mathcal{P}}$, the algorithm initializes an empty action set $\mathcal{A}$ (line \ref{line1-init}) and exhaustively traverses the power set of action components (line \ref{line1-fullhspace}). For each candidate action model $h$, the algorithm first verifies its validity (lines \ref{line1-actionmodel}–\ref{line1-hspace}). It excludes models based on domain-independent constraints (e.g., $add \cap del \neq \emptyset$) or domain-dependent constraints (e.g., mutually exclusive preconditions), though the latter typically require extensive expert knowledge. Even when restricted to domain-independent constraints, exploring the model space entails an exponential complexity of $O(2^{3n})$. The algorithm then retrieves a control policy $\pi \in \Pi_\mathcal{P}$ (line \ref{line1-pispace}) and verifies its consistency with $h$ (line \ref{line1-policy}). Upon a successful match, it instantiates a consistent action $a=\langle h,\pi \rangle$  (line \ref{line1-action}) and appends it to $\mathcal{A}$. Finally, the algorithm induces the condition set $\mathcal{C}$ from the union of all atomic conditions in $\mathcal{A}$ (line \ref{line1-condition}) and returns the resulting BT system $\Phi$. While exhaustive and correct, this algorithm faces significant limitations: (1) the exponential complexity of exploring $\mathcal{H}_\mathcal{P}$, and (2) the practical difficulty of designing $\Pi_\mathcal{P}$ and verifying policy consistency. Notably, automatically synthesizing low-level control policies to achieve specific effects remains a fundamental challenge in robotics \cite{kumar2023robohive}.


To overcome these limitations, we propose CABTO, a principled framework designed for efficient BT grounding. As illustrated in Algorithm \ref{alg-cabto}, CABTO decomposes the grounding process into three phases, leveraging multi-modal contexts to circumvent exhaustive search. The following sections detail the implementation and context acquisition strategies employed in each phase.



\subsubsection{High-level Model Proposal}

CABTO initially initializes the grounded action set $\mathcal{A}$ as empty (Line \ref{line1-initaction}) and defines the unexplored model space $\mathcal{H}_U$ as the complete set of potential action models $\mathcal{H}_{\mathcal{P}}$ (Line \ref{line2-initspaces}). The process commences with an initial proposal phase, where the LLM receives a structured textual prompt defining the task set $\mathcal{P}$. This context encapsulates goal states and initial conditions formalized as first-order logic propositions, alongside the semantics descriptions of scene objects. Leveraging this task-specific context, the LLM identifies a subset of promising models $\mathcal{H}_E$ from $\mathcal{H}_{\mathcal{P}}$ by specifying their symbolic preconditions and effects in a programmatic format (Line \ref{line3-initproposal}):
\begin{equation}\label{equ-initproposal}
    \mathcal{H}_E = \text{\constant{LLM}}(\mathcal{P},\mathcal{H}_\mathcal{P})
\end{equation} 
Empirical results demonstrate that for simple task sets, this initial proposal phase often yields sufficient action models to satisfy the majority of requirements in $\mathcal{P}$.

To accommodate complex scenarios where initial proposals may be incomplete, CABTO employs a refinement loop that iterates until the task set $\mathcal{P}$ is verified as fully solvable using the validated grounded actions in $\mathcal{A}$ (Line \ref{line2-whileHU}). Within this loop, the algorithm assesses the completeness of the current candidate set $\mathcal{H}_E$ by attempting to synthesize BTs for all tasks in $\mathcal{P}$ through BT Planning. Any planning failure triggers the aggregation of diagnostic data into a failure set $\mathcal{I}_{fail}$ (Line \ref{line4-planning}). Each entry $\mathcal{I}_p \in \mathcal{I}_{fail}$ encapsulates critical diagnostics, such as the topological sketches of incomplete BTs and the count of expanded conditions. These metrics provide essential semantic cues, aiding the LLM in identifying symbolic gaps to propose more promising action models.




\begin{equation}
\label{equ-modelproposal}\mathcal{H}' \leftarrow \text{\constant{LLM}}(\mathcal{P},\mathcal{H}_U, \mathcal{I}_{fail})
\end{equation}

Subsequently, proposed models are transferred from $\mathcal{H}_U$ to the candidate set $\mathcal{H}_E$ (Line \ref{line6-deleteh}). This heuristic search iterates until the task set $\mathcal{P}$ is logically spanned by a complete BT system.

\subsubsection{Low-Level Policy Sampling}

This phase verifies the physical consistency of candidate action models $h \in \mathcal{H}_E$ (Line \ref{line-phase2-start}). For each model $h$, we initialize the trial counter $n=0$ and  the policy $\pi$ as null. To bridge the gap between abstract symbolic reasoning and precise physical execution, we propose a hierarchical framework that integrates Molmo\cite{deitke2025molmo} with programmatic code generation. Specifically, within a budget of $N_{max}$ attempts (Line \ref{line-nmax}), the VLM is a programmatic sampler (Line \ref{line-vlm-sample}) that translates high-level semantic intentions into grounded control policies:
\begin{equation}
\label{equ-policysampling_v3} \pi \leftarrow \text{VLM}(h, \Pi_{\mathcal{P}}, \mathcal{I}_e)
\end{equation}
where $\Pi_{\mathcal{P}}$ represents the set of available control interfaces. These interfaces comprise Molmo-based \cite{deitke2024molmo} perception APIs for extracting environmental keypoints, cuRobo-based \cite{2023curobo} (a 7-DoF IK solver) motion control APIs for the robotic arm, and gripper actuation commands. The execution context $\mathcal{I}_e$ serves as a critical nexus for closed-loop iterative refinement. It encapsulates multi-modal diagnostic data, including egocentric visual observations, previously synthesized control code, post-hoc visual feedback, and categorical success/failure signals. By maintaining this high-fidelity temporal trace, the VLM can effectively anchor its subsequent sampling within the physical constraints evidenced by prior execution attempts.


Specifically, the VLM selectively invokes Molmo-based perception tools conditioned on the logical semantics of $h$. When precise spatial grounding is necessitated, the VLM leverages Molmo to extract functional affordances and task-relevant keypoints—such as optimal grasp points or target placement coordinates—directly from visual observation $\mathcal{V}$. Subsequently, the VLM synthesizes these grounded keypoints and parameterized APIs into executable Pythonic code that instantiates the specific control policy $\pi$.

To validate the policy $\pi$, we initialize a simulation $s_0$ such that the initial state satisfies the precondition $pre(h)$ (Line \ref{line-sample-s0}). After execution (Line \ref{line-execute}), we check if the terminal state $s_t$ achieves the expected symbolic effects: $s_t \supseteq (pre(h) \cup add(h) \setminus del(h))$ (Line \ref{line-check-logic}). Upon verification, the grounded action $\langle h, \pi \rangle$ is appended to the action set $\mathcal{A}$.

\begin{table*}[t]
\centering
\setlength{\tabcolsep}{4pt}
\small
\begin{tabular}{llccccccccc}
\toprule
\multirow{3}{*}{\textbf{Robot}} & \multirow{3}{*}{\textbf{Task Set}} & \multicolumn{3}{c}{\textbf{Task Attributes}} & \multicolumn{3}{c}{\textbf{GPT-3.5-Turbo}} & \multicolumn{3}{c}{\textbf{GPT-4o}} \\
\cmidrule(lr){3-5} \cmidrule(lr){6-8} \cmidrule(lr){9-11}
& & \textbf{Acts} & \textbf{Conds} & \textbf{Steps} & \textbf{ASR(w/o $\rightarrow$ w)} & \textbf{CSR(w/o $\rightarrow$ w)} & \textbf{FC} & \textbf{ASR(w/o $\rightarrow$ w)} & \textbf{CSR(w/o $\rightarrow$ w)} & \textbf{FC}\\
\midrule
\multirow{2}{*}{\textbf{Franka}} & \textbf{Cover} & 2.0 & 4.4 & 4.0 & 60.0\% $\rightarrow$ \textbf{66.7\%} & 40\% $\rightarrow$ \textbf{50\%} & 1.6 & 100.0\% $\rightarrow$ \textbf{100.0\%} & 100\% $\rightarrow$ \textbf{100\%} & 0.0\\
& \textbf{Blocks} & 2.3 & 3.1 & 4.1 & 70.0\% $\rightarrow$ \textbf{70.0\%} & 30\% $\rightarrow$ \textbf{50\%} & 2.0 & 60.0\% $\rightarrow$ \textbf{80.0\%} & 50\% $\rightarrow$ \textbf{80\%} & 1.1\\
\midrule
\multirow{3}{*}{\makecell{\textbf{Dual-}\\\textbf{Franka}}} & \textbf{Pour} & 5.5 & 8.1 & 3.6 & 80.0\% $\rightarrow$ \textbf{96.7\%} & 70\% $\rightarrow$ \textbf{90\%} & 0.5 & 66.7\% $\rightarrow$ \textbf{100.0\%} & 60\% $\rightarrow$ \textbf{100\%} & 0.6\\
& \textbf{Handover} & 5.0 & 3.3 & 2.7 & 80.0\% $\rightarrow$ \textbf{90.0\%} & 70\% $\rightarrow$ \textbf{90\%} & 0.5 & 56.7\% $\rightarrow$ \textbf{90.0\%} & 30\% $\rightarrow$ \textbf{90\%} & 1.3\\
& \textbf{Storage} & 6.0 & 5.4 & 2.2 & 56.7\% $\rightarrow$ \textbf{73.3\%} & 0\% $\rightarrow$ \textbf{60\%} & 2.0 & 53.3\% $\rightarrow$ \textbf{76.7\%} & 20\% $\rightarrow$ \textbf{70\%} & 1.7\\
\midrule
\multirow{2}{*}{\textbf{Fetch}} & \textbf{Tidy Home} & 5.6 & 6.0 & 2.9 & 53.3\% $\rightarrow$ \textbf{56.7\%} & 40\% $\rightarrow$ \textbf{50\%} & 0.7 & 53.3\% $\rightarrow$ \textbf{90.0\%} & 30\% $\rightarrow$ \textbf{90\%} & 1.3\\
& \textbf{Cook Meal} & 6.8 & 7.9 & 5.1 & 70.0\% $\rightarrow$ \textbf{70.0\%} & 50\% $\rightarrow$ \textbf{60\%} & 0.7 & 73.3\% $\rightarrow$ \textbf{100.0\%} & 60\% $\rightarrow$ \textbf{100\%} & 0.4\\
\midrule
\multicolumn{2}{l}{\textbf{Total}} & 4.7 & 5.5 & 3.5 & 67.1\% $\rightarrow$ \textbf{74.8\%} & 42.9\% $\rightarrow$ \textbf{64.3\%} & 1.1 & 66.2\% $\rightarrow$ \textbf{91.0\%} & 50\% $\rightarrow$ \textbf{90.0\%} & 0.9\\
\bottomrule
\end{tabular}
\caption{High-level model proposal results (averaged over 10 trials, max FC=3) for GPT-3.5-turbo vs.\ GPT-4o. Note: ``w/o" denotes without planning contexts, and ``w" denotes with planning contexts. }

\label{tab:exp1}
\end{table*}

\begin{figure*}
    \centering
    \includegraphics[width=1\linewidth]{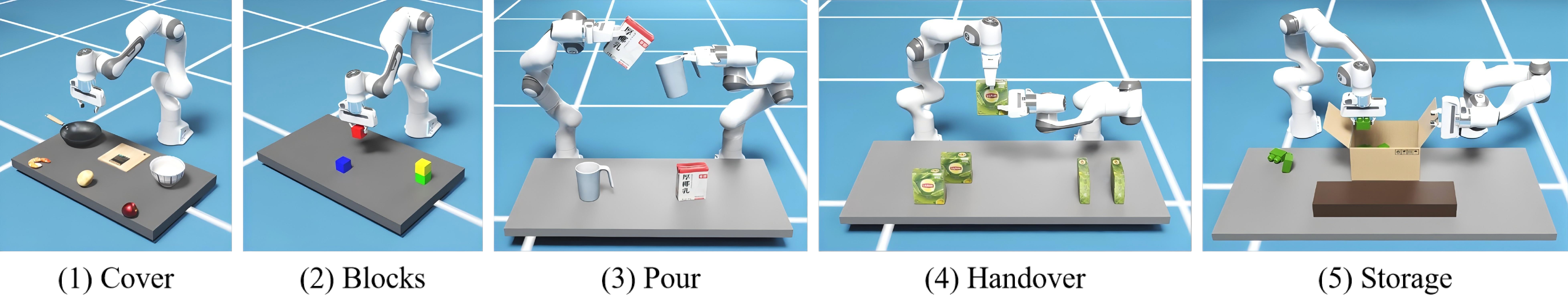}

    \caption{Configurations of the single-arm and dual-arm Franka manipulation tasks in Isaac Sim.}
    \label{fig:franka}
\end{figure*}

\begin{figure*}[h!]
	\centering
	\small
    \includegraphics[width=\textwidth]{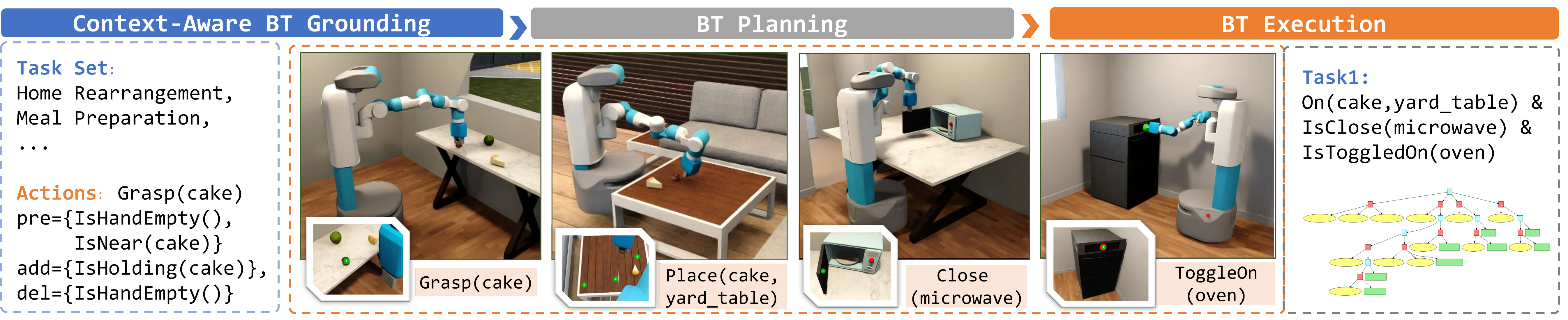}

	\caption{The deployment of CABTO in OmniGibson: Given a task set, CABTO generates a complete and consistent BT system. For a specific task, BT planning is used to generate the solution BT. Then the BT is executed, enabling the robot to successfully achieve the goal.}
	\label{fig:development}
\end{figure*}

\subsubsection{Cross-Level Refinement}

If a sufficient number of policies fails to yield a valid policy, the action model $h$ is deemed physically inconsistent. While a naive approach would be to discard the model and restart the high-level proposal in the next iteration, we instead leverage both planning and execution contexts to refine $h$ (Line \ref{line2-crosslevel}): 
\begin{equation}\label{equ-crosslevel}
h'  \leftarrow \text{\constant{VLM}}(h,\mathcal{H}_U,\left\{\mathcal{I}_p\right\},\Pi_\mathcal{P},\left\{\mathcal{I}_e\right\}) 
\end{equation}
Here, the VLM synthesizes the planning context $\{\mathcal{I}_p\}$, which defines the functional necessity of $h$ within successful symbolic sequences ($\forall \mathcal{I}_p \in \{\mathcal{I}_p\}, h \in \mathcal{T}_p$), and the execution context $\{\mathcal{I}_e\}$, which comprises multi-modal diagnostic data such as egocentric pre/post-action imagery and binary feedback. By integrating these cross-level insights, the VLM identifies underlying failures—such as omitted spatial preconditions or inaccurate symbolic effects—to synthesize a rectified action model $h'$.

Upon completing the refinement loop, a knowledge synchronization step (Line \ref{line32-updateh}) updates $\mathcal{H}_E \leftarrow \mathcal{H}$, ensuring the explored pool consists exclusively of models verified by physical policies. This update provides a grounded, reliable action library for subsequent planning iterations (Line \ref{line4-planning}). The cycle repeats until a set of grounded actions $\mathcal{A}$ renders all tasks $\mathcal{P}$ solvable, after which the condition set $\mathcal{C}$ is extracted to define the final grounded state space (Line \ref{line2-conditionset}).


\begin{table*}[t]
\centering
\setlength{\tabcolsep}{6pt} 
\small
\begin{tabular}{lccccc|cc}
\toprule
 \multirow{3}{*}{\textbf{Action}} & \textbf{\makecell{End-to-end}} & \multicolumn{3}{c}{\textbf{Hierarchical}} & \textbf{\makecell{Rule-based}} & \multicolumn{2}{c}{\textbf{Molmo+cuRobo+APIs}} \\  

\cmidrule(lr){2-2} \cmidrule(lr){3-5} \cmidrule(lr){6-6} \cmidrule(lr){7-8}
 
 & \textbf{\makecell{OpenVLA}} & \textbf{\makecell{VoxPoser}} & \textbf{\makecell{ReKep}} &  \textbf{\makecell{Molmo+cuRobo}} & \textbf{\makecell{APIs}} & \textbf{\makecell{w/o Contexts}} & \textbf{\makecell{with Contexts}}\\
\midrule
$\mathtt{Pick}(obj)$ & 4/10 & 4/10 & 6/10 & 5/10 & 6/10 & 6/10 & \textbf{7/10}\\
$\mathtt{Place}(obj,loc)$ & 5/10  & 3/10 & 7/10 & 5/10 & 6/10 & 6/10 & \textbf{8/10}\\
$\mathtt{Open}(container)$ & 1/10  & 1/10 & 1/10 & 3/10 & 1/10 & 2/10 & \textbf{4/10}\\
$\mathtt{Close}(container$ & 2/10  & 2/10 & 3/10 & 4/10 & 2/10 & 3/10 & \textbf{5/10}\\
$\mathtt{Toggle}(switch)$ & 2/10  & 1/10 & 4/10 & 6/10 & 5/10 & 5/10 & \textbf{7/10}\\
\midrule
\textbf{Total} & 28\% & 22\% & 42\% & 46\% & 40\% & 44\% & \textbf{62\%} \\
\bottomrule
\end{tabular}
\caption{Evaluation results of low-level policy sampling using VLM for 5 typical action models.}
\label{tab:exp2}
\end{table*}

\begin{table*}[h]
    \centering
    \small
    \setlength{\tabcolsep}{4.5pt} 
    \label{tab:vlm_correction_estimation_final_v3}
    \begin{tabular}{l l c c c c}
        \toprule 
        \textbf{Action} & \textbf{Defect Type \& Description} & \makecell{\textbf{Textual} \\ \textbf{Baseline}} & \makecell{\textbf{w/o}  \\ \textbf{Feedback}} & \makecell{\textbf{with}  \\ \textbf{Feedback}} & \makecell{\textbf{Avg. FC}} \\
        \midrule
        $\mathtt{PutIn}(obj, container)$ & \textbf{Pre:} Missing $\mathtt{IsOpen}(container)$ due to closed lid & 10\% & 40\% & \textbf{80\%} & 1.1 \\
        $\mathtt{Stack}(obj\_a, obj\_b)$ & \textbf{Pre:} Missing $\mathtt{Clear}(obj\_b)$ due to surface obstruction & 20\% & 30\% & \textbf{70\%} & 2.1 \\
        $\mathtt{Lift}(box_{big}, r_1, r_2)$ & \textbf{Pre:} Missing $\mathtt{Holding}(r_2, box_{big})$ in dual-arm coordination & 10\% & 80\% & \textbf{90\%} & 0.3 \\
        $\mathtt{Pick}(robot, obj)$ & \textbf{Add:} Unverified $\mathtt{InReach}(robot, obj)$ (Kinematic constraint) & 20\% & 50\% & \textbf{90\%} & 0.8 \\
        $\mathtt{Put}(obj, loc)$ & \textbf{Del:} Stale $\mathtt{At}(obj, loc_{old})$ resulting in location redundancy & 0\% & 20\% & \textbf{40\%} & 2.4 \\
        \midrule
        \multicolumn{2}{l}{\textbf{Total}} & 12\% & 44\% & \textbf{74\%} & 1.3 \\
        \bottomrule
        
    \end{tabular}
    \caption{Success rate (SR\%) of VLM-based cross-level refinement for action models. Results are averaged over 10 trials ($N_{FC} \le 3$). \textbf{Pre}, \textbf{Add}, and \textbf{Del} represent action precondition, add effect, and delete effect, respectively.}
    \label{tab:exp3}
\end{table*}

\subsection{Experimental Setup}

\paragraph{Task Sets}  
We evaluate the robustness and adaptability of CABTO on a comprehensive suite of seven robotic manipulation task sets, encompassing 21 unique goals (three goals per task) across three distinct robotic platforms. These scenarios are strategically designed to cover a spectrum of physical and logical challenges: Single-Arm Franka (T1: Cover, T2: Blocks), Dual-Arm Franka (T3: Pour, T4: Handover, T5: Storage), and Mobile Fetch (T6: Tidy Home, T7: Cook Meal). As summarized in Table \ref{tab:exp1}, these tasks range from fundamental pick-and-place and stacking (T1--T2) to complex bimanual coordination for cooperative transport and exchange (T3--T5), and long-horizon mobile manipulation involving articulated objects and semantic state changes (T6--T7). To quantify solution complexity, Table \ref{tab:exp1} reports the resulting BT attributes for each task set, including the number of unique action predicates ($\text{Acts}$), condition predicates ($\text{Conds}$), and the total execution steps ($\text{Steps}$).

\paragraph{Environment}  
Fetch robot experiments were conducted in OmniGibson \cite{li2023behavior1k} for its realistic physics, while Franka tasks were designed in Isaac Sim to enable flexible object configuration (Figures \ref{fig:franka} and Figures \ref{fig:development}). All experiments are conducted on a single NVIDIA RTX 4090 GPU.

\paragraph{Metrics}

We evaluate the completeness of the high-level model using two primary metrics: (1) Average Planning Success Rate ($\text{ASR}$): The mean planning success rate across all individual tasks within a given task set. (2) Complete Planning Success Rate ($\text{CSR}$): The success rate where all tasks within the set are successfully planned simultaneously. We also report the average number of Feedback Cycles ($\text{FC}$).

\subsection{Evaluation of High-Level Model Proposal}

\paragraph{Ablating Planning Contexts} 
Planning context feedback proved crucial for performance (Table \ref{tab:exp1}). Its inclusion consistently boosted goal success rates and system completeness, most notably for GPT-4o, where completeness jumped from  50\% to over 90\%. The performance gains were most significant in the complex dual-arm and mobile manipulation tasks, demonstrating that structured, symbolic feedback from a formal planner can empower LLMs to resolve intricate logical challenges.



\paragraph{Comparison of LLMs} 
As shown in Table \ref{tab:exp1}, while GPT-3.5 and GPT-4o performed comparably without planning context feedback, GPT-4o's superiority became evident with it. Guided by this feedback, GPT-4o achieved over 90\% complete planning success rate, in stark contrast to approximately 60\% for GPT-3.5. This underscores GPT-4o's advanced capacity for leveraging contextual feedback in complex reasoning tasks like BT grounding.


\subsection{Evaluation of Low-Level Policy Sampling}

We evaluate the performance of three policy types for low-level policy sampling. Details of these policy are shown in Appendix. We select five typical action models to test the performance of these polices, as shown in Table \ref{tab:exp2}. The algorithms show different strengths in various actions. Rekep and Rule-Based methods excel in grasping, while Molmo+cuRobo performs better in $\mathtt{Open/Close}$ and $\mathtt{Toggle}$ actions. This is due to the semantics-based keypoint extraction that accurately identifies object handles and hinges. We utilize GPT-4o as VLM in the experiment.

\paragraph{Ablating Execution Contexts} 
Table \ref{tab:exp2} presents the Success Rate (SR) of control policy for five typical action models, where the VLM samples the policy type and its hyperparameters based on the execution contexts. The results show the SR without execution contexts and with up to three sample attempts. It is evident that the VLM can effectively sample low-level policies, and with execution contexts, the SR of the actions improves.

\subsection{Evaluation of Cross-Level Refinement}
\paragraph{Ablating Environment Feedback}  Table \ref{tab:exp3} catalogs action models that exhibited inconsistencies, where the predicted high-level effect diverged from the low-level execution outcome or resulted in an error. Through an iterative feedback process, the VLM demonstrated the potential to successfully correct these high-level representations, underscoring the critical role of direct environmental feedback. However, the efficacy of this approach is currently limited for abstract concepts lacking direct visual correlates, such as the symbolic target in $\mathtt{Put}(obj, loc)$.

\paragraph{Deployment} Figure \ref{fig:development} depicts the deployment of our pipeline, where CABTO generates a complete and consistent BT system for the given task set. The robot successfully executes the planned BT actions sequentially for every task.

\section{Conclusion}

In this work, we first formalize the BT grounding problem and propose CABTO, a framework that leverages LMs to automatically construct complete and consistent BT systems guided by planning and environmental feedback. The effectiveness of our approach is validated across seven robotic manipulation task sets. Future work will focus on enhancing LM inference and low-level robotic skills via fine-tuning and addressing the transfer to physical systems.

\newpage

\section{Acknowledgments}
This work was supported by the National Science Fund for Distinguished Young Scholars (Grant Nos. 62525213), the National Natural Science Foundation of China (Grant Nos. 62572480), and the University Youth Independent Innovation Science Foundation (Grant Nos. ZK25-11).

\bibliography{aaai2026}

\end{document}